\documentclass[11pt]{article}

\usepackage{times}
\usepackage{geometry}
\usepackage{graphicx}
\usepackage{subcaption}
\usepackage{booktabs}
\usepackage{multirow}
\usepackage{url}
\usepackage{hyperref}
\usepackage{float}
\usepackage{natbib}

\usepackage{amsmath,amsfonts,bm}

\def\eqref#1{equation~\ref{#1}}

\def\1{\bm{1}}

\DeclareMathAlphabet{\mathsfit}{\encodingdefault}{\sfdefault}{m}{sl}
\SetMathAlphabet{\mathsfit}{bold}{\encodingdefault}{\sfdefault}{bx}{n}

\title{ChartDensity-Bench: Benchmarking MLLMs for Numerical Data Reconstruction under Visual Density}

\author{
Xinhe Wu \\
iSoftStone AI Research Lab (ISSAIR)\\
iSoftStone\\
Beijing, China \\
\texttt{xhwude@isoftstone.com}
\and
Yadong Jin\thanks{Corresponding author.}\\
iSoftStone AI Research Lab (ISSAIR)\\
iSoftStone\\
Beijing, China \\
\texttt{ydjink@isoftstone.com}
}

\begin{document}

\maketitle

\begin{abstract}

Multimodal large language models (MLLMs) offer a promising approach for recovering numerical data from scientific charts, but their ability to reconstruct chart data from visually dense figures remains poorly understood. Existing chart understanding benchmarks primarily evaluate question answering or chart-level reasoning and provide limited support for evaluating structured numerical reconstruction from scientific figures. We introduce \textbf{ChartDensity-Bench}, a benchmark for evaluating MLLMs on structured numerical data reconstruction from compound chart figures under controlled visual density. Built from charts paired with source-level ground-truth data, ChartDensity-Bench systematically varies the number of simultaneously presented charts ($k\in{1,3,6,9}$), enabling controlled evaluation of density-induced degradation. We further propose a multi-dimensional evaluation framework covering structural reliability, reconstruction completeness, parseability, and numerical fidelity. Experiments on five recent MLLMs show that numerical reconstruction generally degrades as visual density increases, while the magnitude of degradation varies substantially across models. Chart-level paired comparisons further show that the same source chart can incur higher reconstruction error when embedded in denser visual contexts. These findings highlight visual density as an important and previously underexplored factor in MLLM chart data reconstruction and provide a systematic benchmark for evaluating model robustness in this setting.

\end{abstract}

\section{Introduction}

Scientific literature contains a vast amount of empirical knowledge encoded in figures and charts. As machine learning is increasingly applied to scientific discovery, researchers in \textit{AI for Science} increasingly rely on published literature as a source of training and evaluation data~\cite{wang2023scientific,MOBARAK2023100523}. However, much numerical information is available only as visualizations, while the underlying data are not released or are provided in heterogeneous formats~\cite{yuan2026charterrecover,he2026exchart}. Recovering these data at scale therefore remains labor-intensive, requiring interpretation of axes, legends, data series, and graphical elements. Automated reconstruction of structured numerical data from scientific figures could substantially reduce this bottleneck.

The emergence of multimodal large language models (MLLMs) provides a promising approach to this problem. Recent MLLMs have demonstrated strong capabilities in document understanding, optical character recognition, and chart reasoning~\cite{achiam2023gpt4,wang2024qwen2vl,liu2023ocrbench,liu2026start}. In principle, an MLLM can directly process a scientific figure and reconstruct its underlying data. However, scientific figures often contain multiple subplots, dense data series, small visual elements, and heterogeneous graphical structures. Providing an entire compound figure may therefore introduce substantial visual interference. Decomposing a compound figure into individual subplots is a natural potential solution, but the effect of visual density on numerical data reconstruction has not been systematically studied.

Existing multimodal benchmarks provide limited support for studying this problem. ChartQA and ChartQA Pro evaluate chart understanding and numerical reasoning, but provide limited coverage of scientific charts and often contain explicitly displayed numerical values that can be recognized through OCR~\cite{masry2022chartqa,masry2025chartqapro}. Their numerical evaluation also focuses primarily on point-wise error, without separately characterizing chart separation or reconstruction completeness. ExChart-Bench provides a more direct chart-to-data extraction setting, but its ground truth is manually annotated from chart images and its evaluation primarily emphasizes format success and numerical error~\cite{he2026exchart}. These limitations motivate a benchmark with source-level ground truth and explicit evaluation of structural reliability, completeness, and numerical fidelity.

To address this gap, we introduce \textbf{ChartDensity-Bench}, a benchmark for evaluating MLLMs on structured numerical data reconstruction from compound chart figures under controlled visual density. Rather than answering individual questions about a chart, models must identify each constituent chart and reconstruct its associated tabular data, including data-series identifiers, $x$-axis entries, and numerical values. The benchmark is constructed from charts paired with ground-truth tables, enabling direct evaluation against the underlying source data.

A key feature of ChartDensity-Bench is its controlled visual-density setting. We systematically compose $k\in{1,3,6,9}$ source charts into compound figures, allowing the same source charts to be evaluated under different numbers of simultaneously presented subplots. We evaluate reconstruction from complementary perspectives: \textit{structural reliability}, \textit{reconstruction completeness}, and \textit{numerical fidelity}, while additionally reporting \textit{parseability} to distinguish formatting or structural failures from numerical errors. Across five recent MLLMs, we find that reconstruction quality generally degrades as visual density increases, with substantial differences in density sensitivity across models.

Our contributions are summarized as follows:
\begin{itemize}
\item We introduce \textbf{ChartDensity-Bench}, a benchmark for structured numerical data reconstruction from compound scientific chart figures with source-level ground truth under controlled visual density.
\item We propose a \textbf{multi-dimensional evaluation framework} covering structural reliability, reconstruction completeness, parseability, and numerical fidelity.
\item We systematically evaluate five recent MLLMs under controlled compound-figure settings and characterize their differing sensitivity to increasing visual density.
\end{itemize}

\section{ChartDensity-Bench}

We introduce \textbf{ChartDensity-Bench}, a benchmark for evaluating the
ability of multimodal large language models (MLLMs) to reconstruct
structured numerical data from compound chart figures under
controlled visual density. Unlike conventional chart
understanding benchmarks that primarily evaluate visual question
answering or semantic interpretation, ChartDensity-Bench requires models to
recover the underlying tabular data of every constituent chart.
This formulation exposes three distinct failure modes:
\textit{structural reliability}, \textit{reconstruction completeness},
and \textit{numerical fidelity}.

\subsection{Benchmark Overview}

Given a compound figure containing $k$ charts, where
$k\in\{1,3,6,9\}$, an MLLM is required to identify the constituent
charts and reconstruct the data table associated with each chart.
Each table contains data-series identifiers, x-axis entries, and
their corresponding numerical values. Importantly, the model is not
provided with chart-level annotations or source-chart identifiers.

The task therefore requires three levels of correspondence:
(i) \textit{structural correspondence}, i.e., correctly separating
the constituent charts; (ii) \textit{point correspondence}, i.e.,
matching extracted series and x-axis entries to the ground truth;
and (iii) \textit{numerical correspondence}, i.e., accurately
recovering the underlying y-values. We evaluate these capabilities
under four controlled visual-density levels, with $k=1$ serving as
the single-chart baseline and larger $k$ corresponding to increasingly
dense compound figures.

\begin{figure*}[t]
    \centering
    \includegraphics[width=0.8\textwidth]{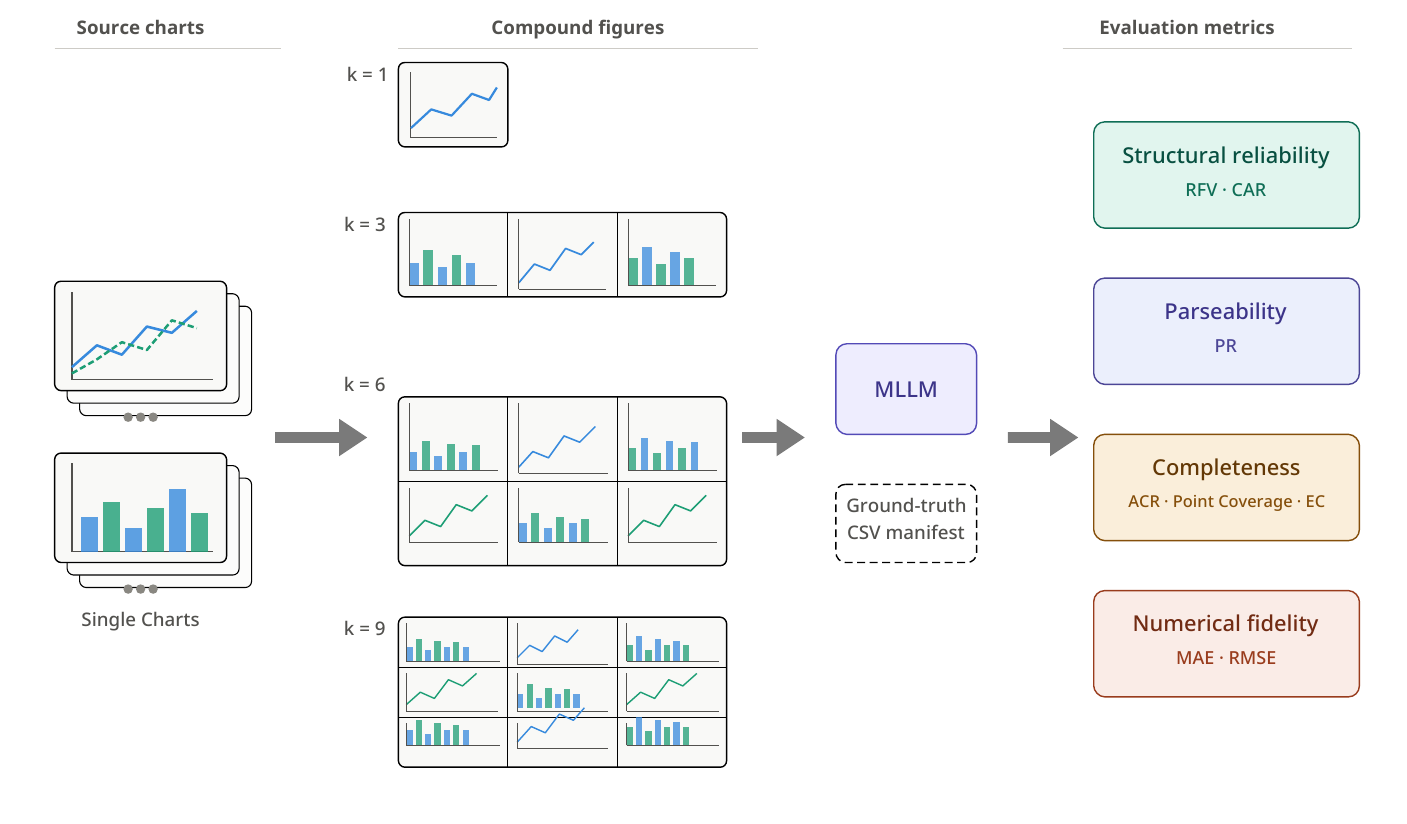}
    \caption{
    Overview of ChartDensity-Bench. Source charts are systematically composed
    into increasingly dense compound figures with $k\in\{1,3,6,9\}$
    charts. MLLMs must identify each constituent chart and reconstruct
    its underlying numerical data. The benchmark separately measures
    structural reliability, reconstruction completeness, and numerical
    fidelity, enabling controlled analysis of performance degradation
    under increasing visual density.
    }
    \label{fig:overview}
\end{figure*}

\subsection{Dataset Construction}

ChartDensity-Bench is constructed from the \textit{ICPR CHART-Infographics}
dataset~\cite{davila2022icpr}, which provides chart images paired with ground-truth
tables. We focus on line and bar charts and systematically compose
individual charts into compound figures.

\paragraph{Controlled visual density.}
For each density level $k\in\{1,3,6,9\}$, we compose $k$ source
charts into a single image. The $k=1$ condition uses the original
chart, while $k=3$, $6$, and $9$ arrange charts in $1\times3$,
$2\times3$, and $3\times3$ grids, respectively. Adjacent charts are
separated by a fixed 10-pixel margin. Each source image is resized
to a common width of 800 pixels while preserving its aspect ratio.
Thus, the number of simultaneously presented charts provides a
controlled density variable while preserving the underlying chart
content.

\paragraph{Ground-truth manifest.}
For every compound figure, we retain the ordered list of source-chart
identifiers and their corresponding ground-truth. This
manifest provides the reference for chart-level and point-level
alignment during evaluation.

\paragraph{Data partitioning.}
We use a fixed random seed of 42 for reproducible composition. Source
charts are sampled without replacement within each compound, and
incomplete groups are discarded. The resulting benchmark contains
four controlled density conditions corresponding to $1$, $3$, $6$,
and $9$ simultaneously presented charts.

\subsection{Data Reconstruction and Alignment}

Each compound figure is independently provided to the target MLLM
with a unified extraction prompt. The model is instructed to identify
individual charts, distinguish data series, interpret x-axis values,
and recover chart-specific data points. For line charts, values are
associated with visible line markers or nodes; for bar charts, values
are associated with the centers of bar tops. The model is required
to produce a standardized table representation:

\begin{center}
\texttt{| Series Name | X Value | Y Value |}.
\end{center}

The raw responses are subsequently parsed into chart-level tables.
Explicit chart delimiters are used when available, while structural
cues such as x-axis sequences and series structure are used for
continuous outputs. Formatting and parsing failures are recorded
separately from numerical reconstruction errors.

Predicted tables are then matched to the source charts using the
ground-truth manifest. Within each chart, series are matched using
normalized string similarity with one-to-one assignment. Numerical
x-values are matched by nearest-neighbor assignment subject to a
predefined tolerance, while categorical x-values are matched by
their labels or source ordering. Only successfully matched points
are used for numerical fidelity evaluation.

\subsection{Evaluation Metrics}

We evaluate reconstruction quality along four complementary dimensions: \textbf{structural reliability}, \textbf{parseability}, \textbf{completeness}, and \textbf{numerical fidelity}. We report all metrics separately for visual densities $k\in{1,3,6,9}$. Detailed definitions and implementation details are provided in Appendix~\ref{app:detailed_metrics}.

\paragraph{Structural reliability.}
We report \textbf{Response Format Validity (RFV)}, which measures whether the predicted number of charts matches the expected number at the compound level, and \textbf{Chart Alignment Rate (CAR)}, which measures the fraction of expected charts that are recovered from compounds with structurally valid chart counts. Thus, RFV captures compound-level structural correctness, while CAR measures chart-level recovery.

\paragraph{Parseability.}
We report \textbf{Parseability Rate (PR)}, the fraction of aligned charts for which valid numerical evaluation can be performed. A chart is considered parseable when valid point correspondences can be established and numerical errors can be computed. Charts for which no valid matches remain, or whose resulting MAE or RMSE exceeds 100, are treated as unparseable. These cases predominantly arise from unsuccessful structural matching, such as incorrect row ordering or series assignment.

\paragraph{Completeness.}
For each aligned chart $i$, we define the \textbf{Completeness Ratio} as
\begin{equation}
C_i=\frac{\hat N_i}{N_i},
\end{equation}
where $N_i$ and $\hat N_i$ denote the numbers of valid ground-truth and extracted $x$-axis entries, respectively. We report the mean across aligned charts as \textbf{Average Completeness Ratio (ACR)}. We additionally report \textbf{Point Coverage}, which measures the fraction of valid ground-truth points that are successfully matched by the extracted reconstruction.

To jointly account for chart-level recovery and within-chart completeness, we define \textbf{Effective Completeness (EC)} as
\begin{equation}
\mathrm{EC}=\mathrm{CAR}\times\mathrm{ACR}.
\end{equation}

\paragraph{Numerical fidelity.}
For successfully matched point pairs, we normalize the extracted and ground-truth $y$-values to $[0,100]$ using the minimum and maximum ground-truth values among the matched points within the corresponding chart. We then report \textbf{Mean Absolute Error (MAE)} and \textbf{Root Mean Squared Error (RMSE)}. When normalization is not possible, relative errors are used as a fallback, followed by bounded absolute errors.

For multi-series charts, we first attempt to associate extracted and ground-truth series using normalized string matching based on edit distance~\cite{levenshtein1966binary}. If a reliable series mapping cannot be established, we fall back to sequential row-order matching. Within successfully mapped series, numeric $x$-values are matched using one-to-one nearest-neighbor assignment~\cite{kuhn1955hungarian}, while categorical values are matched by row order. Further details are provided in Appendix~\ref{app:detailed_metrics}.

\section{Benchmark Results}
\label{sec:benchmark_results} 
We evaluate five API-accessed MLLMs---Doubao-Seed-2.0-Pro~\cite{bytedance2026doubao},
Gemini-3.1-Pro~\cite{google2026gemini31}, GLM-4.6V~\cite{glmvteam2025glm}, Kimi-K2.6~\cite{moonshotai2025kimik2}, and Qwen-3.6-Plus~\cite{bai2025qwen3vl}---under four
controlled visual-density conditions with 1, 3, 6, and 9 simultaneously
presented charts. We ask three questions: (i) how accurately do current
MLLMs reconstruct chart data, (ii) how does reconstruction quality
change as visual density increases, and (iii) do different models exhibit
distinct sensitivity to dense visual contexts?

\subsection{Overall Reconstruction Performance}

Table~\ref{tab:main_results} summarizes reconstruction performance
across density levels. Even in the single-chart setting, where no
competing charts are present, substantial differences exist across
models. Gemini achieves the strongest overall reconstruction quality,
obtaining both the highest point coverage (60.5\%) and the lowest
normalized MAE (5.51), followed by Qwen, Kimi, Doubao, and GLM in
terms of numerical error. This result indicates that chart data
reconstruction remains non-trivial even without additional visual
interference.

More importantly, single-chart performance does not fully predict
robustness to dense compound figures. While Gemini provides the best
single-chart numerical accuracy, Kimi exhibits a substantially smaller
increase in error as the number of simultaneously presented charts
grows. This distinction between absolute reconstruction quality and
density robustness motivates a direct analysis of performance
degradation under increasing visual density.

\subsection{Effect of Increasing Visual Density}

\begin{figure}[t]
    \centering
    \begin{subfigure}[t]{0.49\linewidth}
        \centering
        \includegraphics[width=\linewidth]{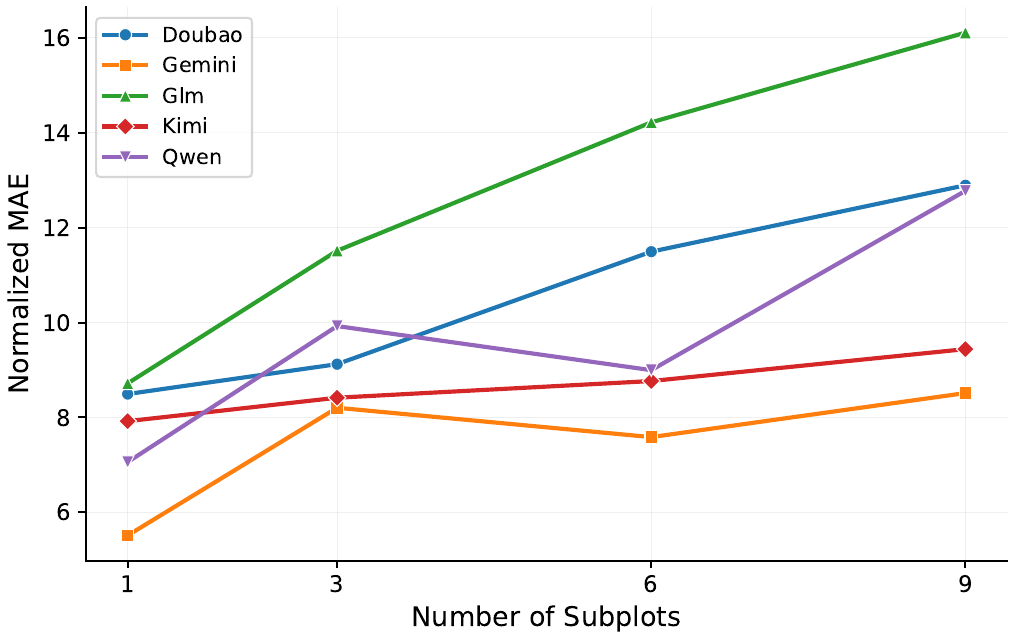}
        \caption{}
        \label{fig:mae_density}
    \end{subfigure}
    \hfill
    \begin{subfigure}[t]{0.49\linewidth}
        \centering
        \includegraphics[width=\linewidth]{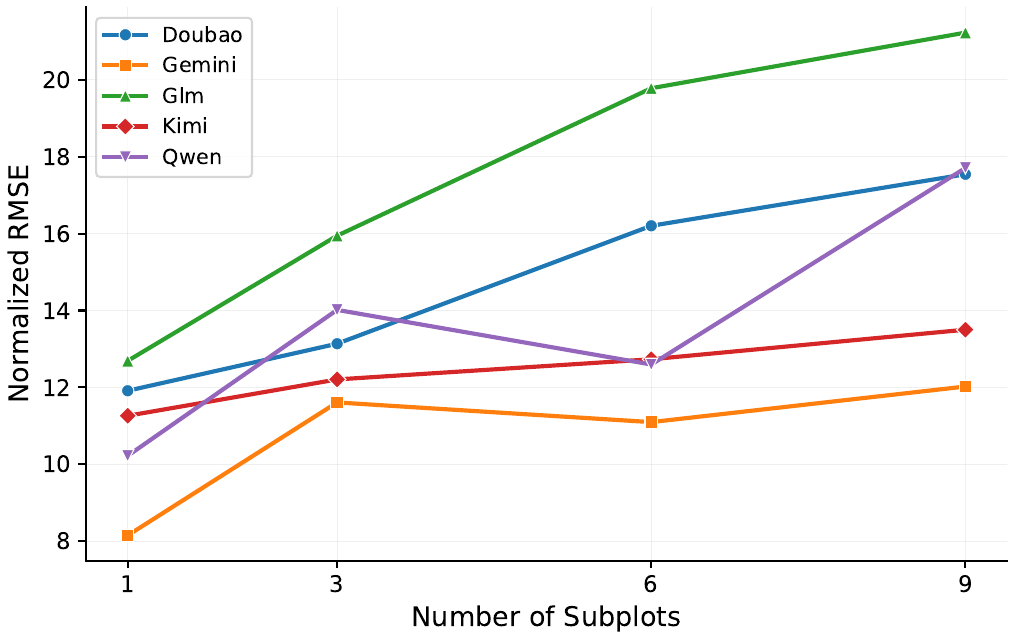}
        \caption{}
        \label{fig:rmse_density}
    \end{subfigure}
    \caption{
    Numerical reconstruction error under increasing chart density.
    The x-axis denotes the number of simultaneously presented subplots.
    Lower values indicate better reconstruction fidelity.
    }
    \label{fig:density_error}
\end{figure}

\begin{table}[t]
\centering
\small
\setlength{\tabcolsep}{4pt}
\caption{
Density-induced degradation in numerical reconstruction from 1 to 9 subplots.
$\Delta$ denotes the absolute increase in error, while Relative $\Delta$ measures the increase relative to the single-chart baseline.
}
\label{tab:density_delta}
\begin{tabular}{lrrrr}
\toprule
Model & $\Delta$MAE $\uparrow$ & Rel. $\Delta$MAE $\uparrow$ &
$\Delta$RMSE $\uparrow$ & Rel. $\Delta$RMSE $\uparrow$ \\
\midrule
Doubao & 4.4009 & 51.82\% & 5.6338 & 47.30\% \\
Gemini & 3.0017 & 54.46\% & 3.8754 & 47.57\% \\
GLM    & 7.3937 & 84.81\% & 8.5409 & 67.31\% \\
Kimi   & 1.5191 & 19.18\% & 2.2397 & 19.89\% \\
Qwen   & 5.7183 & 81.01\% & 7.4899 & 73.30\% \\
\bottomrule
\end{tabular}
\end{table}

Figure~\ref{fig:density_error} shows normalized MAE and RMSE as the number of simultaneously presented charts increases from 1 to 9. Reconstruction error generally increases with visual density across all five models, although the trajectories are not strictly monotonic. The effect is strongest for GLM and weakest for Kimi, while Gemini and Qwen show intermediate sensitivity.

The 1-to-9 degradation in Table~\ref{tab:density_delta} further quantifies this model-dependent effect. Relative MAE degradation ranges from 19.18\% for Kimi to 84.81\% for GLM, with RMSE showing the same overall ordering. These consistent trends across two numerical error measures indicate that density-induced degradation is not specific to a single fidelity metric.

\subsection{Completeness--Fidelity Trade-off}

\begin{figure}[t]
    \centering
    \begin{subfigure}[t]{0.49\linewidth}
        \centering
        \includegraphics[width=\linewidth]{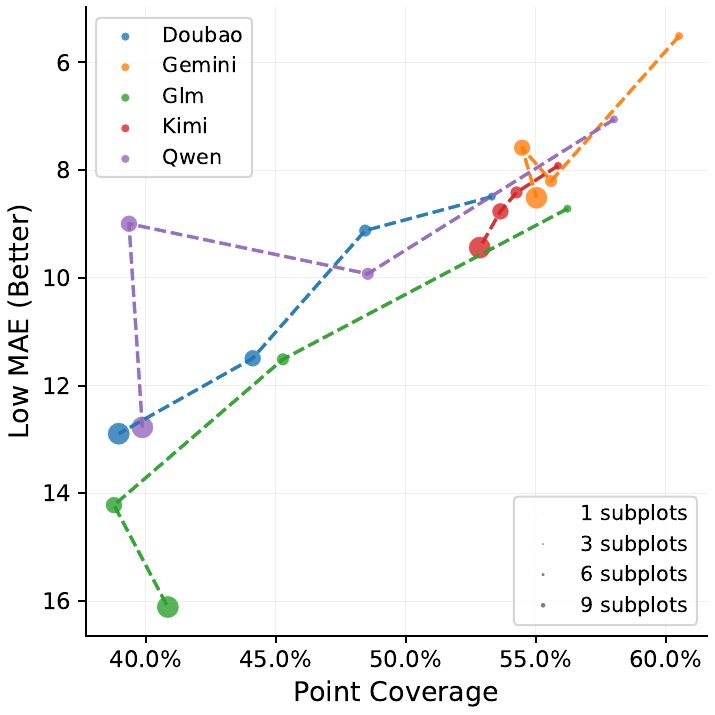}
        \caption{}
        \label{fig:coverage_mae}
    \end{subfigure}
    \hfill
    \begin{subfigure}[t]{0.49\linewidth}
        \centering
        \includegraphics[width=\linewidth]{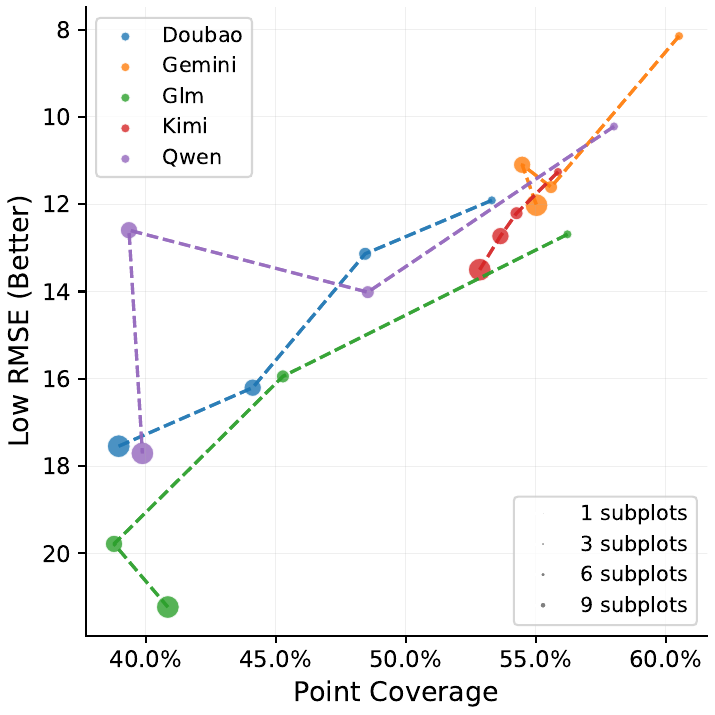}
        \caption{}
        \label{fig:coverage_rmse}
    \end{subfigure}
    \caption{
    Relationship between point coverage and numerical reconstruction
    error under increasing chart density. Better reconstruction lies toward the upper-right region, corresponding to higher point coverage and lower numerical error.
    }
    \label{fig:coverage}
\end{figure}

Numerical accuracy alone does not capture whether a model reconstructs
the complete chart data. Figure~\ref{fig:coverage} therefore jointly
visualizes point coverage and numerical error under increasing density.
Each trajectory connects the four density conditions for the same model.

Overall, denser compound figures tend to move models toward lower point
coverage and higher numerical error, indicating that visual density
induces both omission and value-reconstruction errors. However, the two
failure modes are not perfectly coupled.

Kimi provides a robustness-oriented example: its point coverage
decreases only moderately from 55.9\% to 52.9\% between one and nine
charts, while its MAE increases from 7.92 to 9.44. In contrast, GLM
experiences simultaneous degradation in both dimensions, with point
coverage decreasing from 56.2\% to 40.9\% and MAE increasing from
8.72 to 16.11.

Qwen further illustrates that completeness and numerical fidelity can
change independently. Its point coverage continues to decrease from
three to six charts, while its MAE temporarily improves from 9.93 to
9.00. Thus, a model may recover fewer points while estimating the
remaining points more accurately. This observation motivates treating
completeness and numerical fidelity as separate evaluation dimensions.

\subsection{Model-specific Density Sensitivity}

The results reveal substantial differences in how models respond to
increasing visual density. Kimi is the most density-robust model among
the evaluated systems, exhibiting only a 19.18\% relative increase in
MAE from one to nine charts. GLM shows the strongest sensitivity, with
an 84.81\% relative increase. Qwen also exhibits pronounced sensitivity,
with an 81.01\% relative increase in MAE.

Structural reliability shows a similarly model-dependent pattern.
GLM's response format validity decreases from 100\% in the single-chart
condition to 55.3\% at nine charts, whereas Kimi remains above 98\%
throughout. These results suggest that dense visual contexts can affect
reconstruction at multiple levels, ranging from structural parsing to
point recovery and numerical estimation.

Overall, the experiments establish that visual density is a meaningful
and model-dependent source of reconstruction degradation. In particular,
the substantially different responses of models such as Kimi and GLM
raise the question of why some models remain robust while others
degrade sharply. We investigate this question through attention-level
analysis in Section~\ref{sec:attention_dilution}.

\section{Chart-level Analysis of Density-induced Error}
\label{sec:attention_dilution}

The aggregate results in Section~\ref{sec:benchmark_results} show that numerical reconstruction generally deteriorates as more charts are presented simultaneously. However, aggregate comparisons may confound density with differences in chart composition. We therefore exploit the controlled construction of ChartDensity-Bench and compare the reconstruction of the \emph{same source chart} under different density conditions.

\subsection{Paired Chart-level Density Effects}

\begin{figure}[t]
\centering
\begin{subfigure}[t]{0.4\linewidth}
\centering
\includegraphics[width=\linewidth]{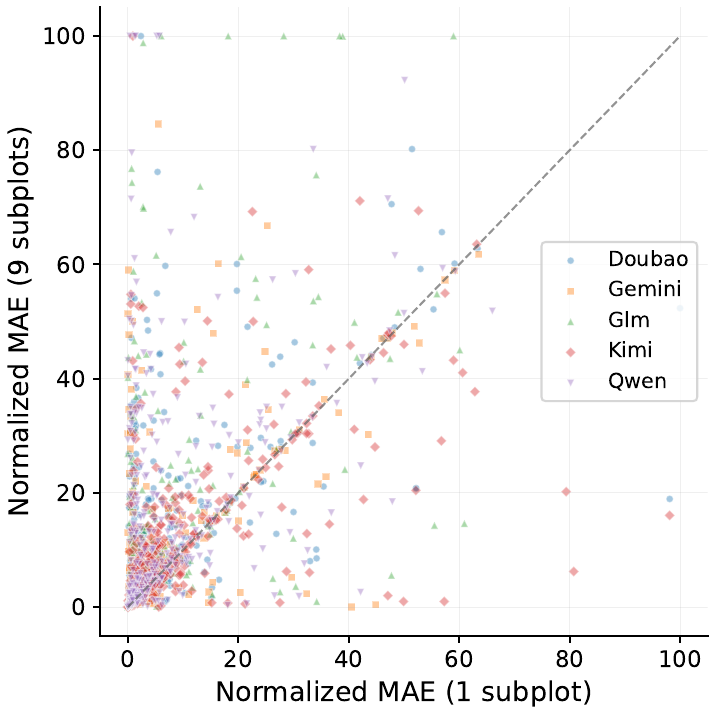}
\caption{}
\label{fig:mae_1vs9}
\end{subfigure}
\hfill
\begin{subfigure}[t]{0.58\linewidth}
\centering
\includegraphics[width=\linewidth]{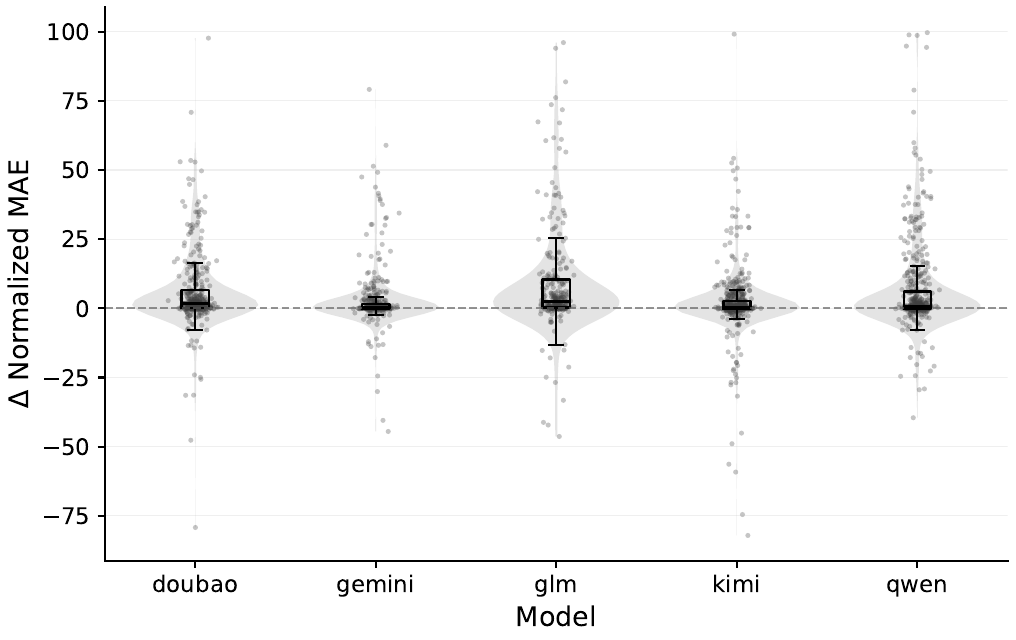}
\caption{}
\label{fig:mae_delta}
\end{subfigure}
\caption{
Paired chart-level comparison between the 1- and 9-chart conditions.
Each point corresponds to the same source chart; values above the diagonal
indicate higher error under density 9. The right panel shows the distribution
of per-chart error changes.
}
\label{fig:per_chart_analysis}
\end{figure}

We first compare reconstruction errors for the same source chart under the single-chart and nine-chart conditions. Figure~\ref{fig:per_chart_analysis} presents paired comparisons of normalized MAE.

A substantial fraction of charts lie above the diagonal in Figure~\ref{fig:per_chart_analysis}a, indicating higher reconstruction error when the same underlying chart is embedded in a nine-chart compound figure. This comparison controls for intrinsic chart difficulty because the source chart is held fixed while its visual context changes.

For source chart $c$, we define the change in reconstruction error as
\begin{equation}
\Delta\mathrm{MAE}_c
=
\mathrm{MAE}_{c,9}-\mathrm{MAE}_{c,1}.
\end{equation}
As shown in Figure~\ref{fig:per_chart_analysis}b, the distribution of $\Delta\mathrm{MAE}$ is generally shifted toward positive values across models, although substantial chart-to-chart variation remains. Thus, density does not affect all charts uniformly, but the aggregate effect is predominantly toward higher reconstruction error. Corresponding RMSE results are provided in Figure~\ref{fig:per_chart_analysis_rmse}.

\subsection{Model-specific Density Sensitivity}

The magnitude of the chart-level density effect varies substantially
across models. To summarize the overall trend, we fit a descriptive
linear trend for each model:

\[
MAE_{c,d} = \alpha_m + \beta_m d + \epsilon_{c,d},
\]

where $d\in\{1,3,6,9\}$ denotes the number of simultaneously presented
subplots. Here, $\beta_m$ summarizes the average direction and magnitude
of the density--error relationship across the four controlled density
levels. Because the same source charts are observed repeatedly, these
slopes are treated as descriptive summaries rather than independent
causal estimates.

\begin{table}[t]
\centering
\small
\setlength{\tabcolsep}{4pt}
\caption{
Per-model sensitivity of chart reconstruction error to visual density.
$\beta$ is the slope from $MAE=\alpha+\beta\cdot Density+\epsilon$.
Positive slopes indicate increasing numerical error as more charts are
presented simultaneously.
}
\label{tab:density_regression}
\begin{tabular}{lccccc}
\toprule
Model & $\beta$ & Std. Error & 95\% CI & $p$ & $R^2$ \\
\midrule
Doubao & 0.582 & 0.117 & [0.354, 0.811] & $<.001$ & .013 \\
Gemini & 0.296 & 0.099 & [0.103, 0.490] & .003 & .004 \\
GLM    & \textbf{0.940} & 0.138 & [0.671, 1.210] & $<.001$ & .027 \\
Kimi   & 0.181 & 0.096 & [-0.008, 0.369] & .060 & .002 \\
Qwen   & 0.627 & 0.109 & [0.413, 0.841] & $<.001$ & .018 \\
\bottomrule
\end{tabular}
\end{table}

As shown in Table~\ref{tab:density_regression}, all five models have
positive estimated density slopes, although the strength of statistical
evidence varies. GLM exhibits the largest slope ($\beta=0.940$),
followed by Qwen ($0.627$), Doubao ($0.582$), Gemini ($0.296$), and
Kimi ($0.181$). The estimated slope for Kimi is the smallest and does
not reach the conventional $p<.05$ threshold ($p=.060$), indicating
weaker evidence for a systematic linear trend across the four density
levels.

The low $R^2$ values reflect substantial heterogeneity in the intrinsic
difficulty of individual charts. They should therefore not be
interpreted as evidence against a density effect. To account explicitly
for repeated observations of the same source charts and to test
model-specific density effects, we next perform a mixed-effects
analysis.

\subsection{Mixed-effects Analysis of Density Sensitivity}

Because each source chart is evaluated under multiple density
conditions and by multiple models, we use a linear mixed-effects model~\cite{bates2015lme4}
with chart identity as a random intercept:

\[
Y_{c,m,d}
=
\beta_0
+
\beta_{\mathrm{model},m}
+
\beta_{\mathrm{density},d}
+
\beta_{\mathrm{model}\times\mathrm{density},m,d}
+
\gamma_c
+
\epsilon_{c,m,d},
\]

where $Y_{c,m,d}$ denotes the normalized reconstruction error for
source chart $c$, model $m$, and density condition $d$. Kimi and the
single-chart condition are used as reference categories.

The results in Table~\ref{tab:mixed_effects} provide evidence for
substantial model-dependent density sensitivity. Relative to Kimi,
Doubao, GLM, and Qwen exhibit significantly larger density effects at
nine subplots, with interaction coefficients of 3.757
($p<.001$), 6.077 ($p<.001$), and 4.803 ($p<.001$), respectively.
The corresponding interaction for Gemini is smaller and does not reach
conventional significance (1.616, $p=.080$).

The main density effect for Kimi at nine subplots is 1.552 MAE points
($p=.014$). Combining this reference effect with the corresponding
model--density interactions gives estimated density-related increases
of approximately 5.31 MAE points for Doubao, 3.17 for Gemini, 7.63 for
GLM, 1.55 for Kimi, and 6.36 for Qwen at nine subplots relative to the
single-chart reference, with the latter quantities interpreted through
the fitted model coefficients.

\subsection{Evidence Consistent with Attention Dilution}

The paired chart-level analysis and mixed-effects results show that increasing visual density is associated with higher reconstruction error even when the underlying source chart is held fixed, with substantially stronger effects for models such as GLM and Qwen than for Kimi. This pattern is consistent with an attention dilution hypothesis~\cite{chen2024fastv, he2024lookm, leng2025rave}, in which additional charts introduce competing visual elements and reduce the effective focus available to each target chart. However, ChartDensity-Bench measures behavioral reconstruction rather than internal attention allocation and therefore cannot establish this mechanism directly. We interpret the results as behavioral evidence compatible with attention dilution, while direct analysis of multimodal attention or internal representations remains an important direction for future work.

\section{Discussion}

Our results have several implications for evaluating MLLMs for numerical chart data reconstruction.

\paragraph{Visual density as an evaluation factor.}
A central finding of ChartDensity-Bench is that visual density itself can substantially affect reconstruction quality. Numerical error generally increases as more charts are presented simultaneously, and the paired chart-level analysis shows that the same source chart can become harder to reconstruct when embedded in a denser compound figure. This suggests that chart reconstruction difficulty depends not only on the intrinsic complexity of the target chart but also on its surrounding visual context. Since scientific figures frequently contain multiple panels, robustness to visual density is therefore an important consideration for chart-to-data extraction systems.

\paragraph{Multi-dimensional evaluation and model robustness.}
The results also show that completeness and numerical fidelity capture distinct failure modes. A model may omit data points while estimating the remaining values relatively accurately, or experience simultaneous degradation in both dimensions. Structural reliability, completeness, parseability, and numerical fidelity therefore provide complementary views of reconstruction behavior. Moreover, strong single-chart performance does not necessarily imply robustness to dense visual contexts: the evaluated models exhibit substantially different degradation patterns, with Kimi showing relatively small density-induced degradation and GLM substantially larger sensitivity.

\paragraph{Limitations and future directions.}
Our benchmark currently focuses on line and bar charts and therefore does not cover the broader diversity of scientific visualizations. In addition, visual density is operationalized by the number of simultaneously presented charts, which provides a controlled variable but does not capture all forms of visual complexity, such as the number of data series, text density, or visual overlap. Finally, the benchmark measures behavioral outcomes and cannot by itself establish the internal mechanism underlying density-induced degradation. Future work could extend the benchmark to additional visualization types and finer-grained sources of visual interference, while directly investigating model representations or inference strategies that improve robustness to dense compound figures.

\section*{Reproducibility statement}

The benchmark construction procedure, controlled density settings, data reconstruction protocol, alignment procedure, and evaluation metrics are described in Sections~2 and 3 and further detailed in the Appendix. We plan to release the benchmark construction and evaluation code upon publication. The benchmark construction and evaluation code will be publicly available at \url{https://github.com/ISSAIR/ChartDensity-Bench}.

\bibliography{references}

\begin{thebibliography}{}

\bibitem[Achiam et~al., 2023]{achiam2023gpt4}
Achiam, J., Adler, S., Agarwal, S., et~al. (2023).
\newblock {GPT-4} technical report.
\newblock {\em arXiv preprint arXiv:2303.08774}.

\bibitem[Bai et~al., 2025]{bai2025qwen3vl}
Bai, S., Cai, Y., Chen, R., et~al. (2025).
\newblock Qwen3-vl technical report.
\newblock {\em arXiv preprint arXiv:2511.21631}.

\bibitem[Bates et~al., 2015]{bates2015lme4}
Bates, D., M{\"a}chler, M., Bolker, B., and Walker, S. (2015).
\newblock Fitting linear mixed-effects models using {lme4}.
\newblock {\em Journal of Statistical Software}, 67(1):1--48.

\bibitem[{ByteDance}, 2026]{bytedance2026doubao}
{ByteDance} (2026).
\newblock Doubao-seed-2.0.
\newblock \url{https://team.doubao.com/seed2}.

\bibitem[Chen et~al., 2024]{chen2024fastv}
Chen, L., Zhao, H., Liu, T., Bai, S., Lin, J., Zhou, C., and Chang, B. (2024).
\newblock An image is worth 1/2 tokens after layer 2: Plug-and-play inference acceleration for large vision-language models.
\newblock In {\em Proceedings of the European Conference on Computer Vision (ECCV)}.

\bibitem[Davila et~al., 2022]{davila2022icpr}
Davila, K., Xu, F., Ahmed, S., Mendoza, D.~A., Setlur, S., and Govindaraju, V. (2022).
\newblock {ICPR} 2022: Challenge on harvesting raw tables from infographics ({CHART-Infographics}).
\newblock In {\em 2022 26th International Conference on Pattern Recognition (ICPR)}, pages 4995--5001. IEEE.

\bibitem[{GLM-V Team} et~al., 2025]{glmvteam2025glm}
{GLM-V Team}, Hong, W., et~al. (2025).
\newblock {GLM-4.5V} and {GLM-4.1V-Thinking}: Towards versatile multimodal reasoning with scalable reinforcement learning.
\newblock {\em arXiv preprint arXiv:2507.01006}.

\bibitem[{Google DeepMind}, 2026]{google2026gemini31}
{Google DeepMind} (2026).
\newblock Gemini 3.1 pro model card.
\newblock \url{https://deepmind.google/models/model-cards/gemini-3-1-pro/}.

\bibitem[He et~al., 2026]{he2026exchart}
He, Y., Ying, P., Cheng, L., Peng, K., Tian, Y., Deng, D., and Wu, Y. (2026).
\newblock Making multimodal llms reliable chart data extractors: A benchmark and training framework.
\newblock {\em arXiv preprint arXiv:2606.29808}.

\bibitem[He et~al., 2024]{he2024lookm}
He, Z. et~al. (2024).
\newblock {LOOK-M}: Look-once optimization in {KV} cache for efficient multimodal long-context inference.
\newblock {\em arXiv preprint arXiv:2406.18139}.

\bibitem[Kuhn, 1955]{kuhn1955hungarian}
Kuhn, H.~W. (1955).
\newblock The {Hungarian} method for the assignment problem.
\newblock {\em Naval Research Logistics Quarterly}, 2(1--2):83--97.

\bibitem[Leng et~al., 2025]{leng2025rave}
Leng, X. et~al. (2025).
\newblock {RAVE}: Re-allocating visual attention in large multimodal models.
\newblock {\em arXiv preprint arXiv:2605.18359}.

\bibitem[Levenshtein, 1966]{levenshtein1966binary}
Levenshtein, V.~I. (1966).
\newblock Binary codes capable of correcting deletions, insertions, and reversals.
\newblock {\em Soviet Physics Doklady}, 10(8):707--710.

\bibitem[Liu et~al., 2023]{liu2023ocrbench}
Liu, Y., Li, Z., Yang, B., et~al. (2023).
\newblock {OCRBench}: On the hidden mystery of {OCR} in large multimodal models.
\newblock {\em arXiv preprint arXiv:2305.07895}.

\bibitem[Liu et~al., 2026]{liu2026start}
Liu, Z., Gao, X., Niu, F., Gao, Q., Liu, L., and Piramuthu, R. (2026).
\newblock Start: Spatial and textual learning for chart understanding.
\newblock In {\em 2026 IEEE/CVF Winter Conference on Applications of Computer Vision (WACV)}, pages 8146--8156. IEEE.

\bibitem[Masry et~al., 2025]{masry2025chartqapro}
Masry, A., Islam, M.~S., Ahmed, M., Bajaj, A., Kabir, F., Kartha, A., Laskar, M. T.~R., Rahman, M., Rahman, S., Shahmohammadi, M., Thakkar, M., Parvez, M.~R., Hoque, E., and Joty, S. (2025).
\newblock {ChartQAPro}: A more diverse and challenging benchmark for chart question answering.
\newblock In {\em Findings of the Association for Computational Linguistics: ACL 2025}, pages 19123--19151. Association for Computational Linguistics.

\bibitem[Masry et~al., 2022]{masry2022chartqa}
Masry, A., Long, D.~X., Tan, J.~Q., Joty, S., and Hoque, E. (2022).
\newblock {ChartQA}: A benchmark for question answering about charts with visual and logical reasoning.
\newblock In {\em Findings of the Association for Computational Linguistics: ACL 2022}, pages 2263--2279. Association for Computational Linguistics.

\bibitem[Mobarak et~al., 2023]{MOBARAK2023100523}
Mobarak, M.~H., Mimona, M.~A., Islam, M.~A., Hossain, N., Zohura, F.~T., Imtiaz, I., and Rimon, M. I.~H. (2023).
\newblock Scope of machine learning in materials research—a review.
\newblock {\em Applied Surface Science Advances}, 18:100523.

\bibitem[{Moonshot AI}, 2025]{moonshotai2025kimik2}
{Moonshot AI} (2025).
\newblock Kimi k2: A trillion-parameter {MoE} language model.
\newblock \url{https://github.com/MoonshotAI/Kimi-K2}.

\bibitem[Wang et~al., 2023]{wang2023scientific}
Wang, H., Fu, T., Du, Y., Gao, W., Huang, K., Liu, Z., Chandak, P., Liu, S., Van~Katwyk, P., Deac, A., Anandkumar, A., Bergen, K.~J., Gomes, C.~P., Ho, S., Kohli, P., Lasenby, J., Leskovec, J., Liu, T.-Y., Manrai, A.~K., Marks, D.~S., Ramsundar, B., Song, L., Sun, J., Tang, J., Veli{\v{c}}kovi{\'{c}}, P., Welling, M., Zhang, L., Coley, C.~W., Bengio, Y., and Zitnik, M. (2023).
\newblock Scientific discovery in the age of artificial intelligence.
\newblock {\em Nature}, 620:47--60.

\bibitem[Wang et~al., 2024]{wang2024qwen2vl}
Wang, P., Bai, S., Tan, S., Wang, S., Fan, Z., Bai, J., Chen, K., Liu, X., Wang, J., Ge, W., et~al. (2024).
\newblock Qwen2-vl: Enhancing vision-language model's perception of the world at any resolution.
\newblock {\em arXiv preprint arXiv:2409.12191}.

\bibitem[Yuan et~al., 2026]{yuan2026charterrecover}
Yuan, Y. et~al. (2026).
\newblock A deep learning framework for scientific chart data extraction and reconstruction.
\newblock {\em Communications Engineering}.

\end{thebibliography}
\bibliographystyle{apalike}

\appendix
\section{Appendix}
\subsection{Detailed Evaluation Metrics}
\label{app:detailed_metrics}

This section provides the formal definitions and implementation details of the evaluation metrics used in ChartDensity-Bench. Reconstruction quality is evaluated at the compound, chart, and point levels, covering structural reliability, parseability, completeness, and numerical fidelity.

\paragraph{Response Format Validity (RFV).}
Let $N_{\mathrm{processed}}$ denote the number of processed compound figures, and let $N_{\mathrm{mismatch}}$ denote the number of compounds for which the number of extracted charts differs from the expected number. We define Response Format Validity as
\begin{equation}
\mathrm{RFV}
=
1-
\frac{N_{\mathrm{mismatch}}}
{N_{\mathrm{processed}}}.
\end{equation}

A compound is considered structurally valid only when the number of extracted charts exactly matches the number of source charts. If the chart counts differ, the compound is recorded as a structural mismatch and its charts are excluded from subsequent chart-level evaluation.

\paragraph{Chart Alignment Rate (CAR).}
Let $N_{\mathrm{aligned}}$ denote the number of charts evaluated from structurally valid compounds, and let $N_{\mathrm{expected}}$ denote the total number of expected charts across all processed compounds. The Chart Alignment Rate is
\begin{equation}
\mathrm{CAR}
=
\frac{N_{\mathrm{aligned}}}
{N_{\mathrm{expected}}}.
\end{equation}

For a visual density $k$, where each compound contains $k$ source charts,
\begin{equation}
N_{\mathrm{expected}}
=
N_{\mathrm{processed}}\times k.
\end{equation}

Because chart-level evaluation is performed only for compounds with the correct chart count, CAR jointly reflects successful compound-level structural recovery and subsequent chart alignment.

\paragraph{Parseability Rate (PR).}
Let $N_{\mathrm{parseable}}$ denote the number of aligned charts for which valid numerical evaluation can be performed. We define
\begin{equation}
\mathrm{PR}
=
\frac{N_{\mathrm{parseable}}}
{N_{\mathrm{aligned}}}.
\end{equation}

A chart is considered parseable when at least one valid point correspondence can be established and a numerical error value can be computed. If no valid matches or error values remain, the chart is considered unparseable. In addition, charts with aggregate MAE or RMSE greater than 100 are treated as unparseable. The latter criterion serves as a numerical sanity check for cases in which structural matching produces degenerate errors.

\paragraph{Completeness Ratio.}
For aligned chart $i$, let $N_i$ denote the number of valid ground-truth $x$-axis entries and let $\hat{N}_i$ denote the number of extracted entries. The chart-level Completeness Ratio is
\begin{equation}
C_i
=
\frac{\hat{N}_i}
{N_i}.
\end{equation}

For numeric $x$-values, only valid numeric entries are counted. For non-numeric $x$-values, non-empty entries are counted.

\paragraph{Average Completeness Ratio (ACR).}
Let $\mathcal{A}$ denote the set of aligned charts. The Average Completeness Ratio is computed as a macro-average over aligned charts:
\begin{equation}
\mathrm{ACR}
=
\frac{1}{|\mathcal{A}|}
\sum_{i\in\mathcal{A}} C_i.
\end{equation}

Thus, each aligned chart contributes equally to ACR regardless of the number of points or series it contains.

\paragraph{Point Coverage.}
For aligned chart $i$, let $L_{i,j}$ denote the number of successfully matched points in series $j$, and let $N^{\mathrm{points}}_{i,j}$ denote the number of valid ground-truth points in that series. We define chart-level Point Coverage as
\begin{equation}
P_i
=
\frac{
\sum_j L_{i,j}
}{
\sum_j N^{\mathrm{points}}_{i,j}
}.
\end{equation}

We report Point Coverage as the macro-average across aligned charts:
\begin{equation}
\mathrm{PointCoverage}
=
\frac{1}{|\mathcal{A}|}
\sum_{i\in\mathcal{A}} P_i.
\end{equation}

This definition distinguishes Point Coverage from ACR: ACR measures the quantity of extracted $x$-axis entries, whereas Point Coverage measures the fraction of ground-truth points that can be successfully matched.

\paragraph{Effective Completeness (EC).}
To jointly account for chart-level recovery and within-chart completeness, we define
\begin{equation}
\mathrm{EC}
=
\mathrm{CAR}\times\mathrm{ACR}.
\end{equation}

EC therefore penalizes both failures to recover the expected charts and incomplete reconstruction within successfully aligned charts.

\paragraph{Series Matching.}
For multi-series charts, extracted and ground-truth series are first grouped according to their series labels. Labels are normalized using case folding and whitespace stripping. We then compute the Levenshtein edit distance~\cite{levenshtein1966binary} between each extracted series name and candidate ground-truth series names.

For an extracted series $s_e$ and its closest ground-truth candidate $s_g$, the mapping is accepted only when
\begin{equation}
d(s_e,s_g)
\leq
\frac{1}{2}
\max\left(|s_e|,|s_g|\right),
\end{equation}
where $d(\cdot,\cdot)$ denotes Levenshtein distance. If a reliable series mapping cannot be established, the evaluator falls back to sequential row-order matching.

\paragraph{$x$-Value Matching.}
Within successfully mapped series, numeric $x$-values are matched using greedy one-to-one nearest-neighbor assignment. Each extracted point is associated with its closest valid ground-truth $x$-value, and candidate matches are processed in ascending order of distance. Each ground-truth point can be used at most once.

For categorical or otherwise non-numeric $x$-values, points are matched according to row order. Only pairs for which both $x$ and $y$ values can be converted to valid numerical values are included in numerical error computation.

\paragraph{Normalization.}
For each aligned chart, let $\mathcal{M}_i$ denote the set of successfully matched point pairs. When at least two matched points are available, we compute
\begin{equation}
y_{\min}^{(i)}
=
\min_{(\hat{y},y)\in\mathcal{M}_i} y,
\qquad
y_{\max}^{(i)}
=
\max_{(\hat{y},y)\in\mathcal{M}_i} y.
\end{equation}

If
$y_{\max}^{(i)}-y_{\min}^{(i)}>0$, the extracted and ground-truth values are normalized to $[0,100]$ as
\begin{align}
\hat{y}_{\ell}^{\mathrm{norm}}
&=
100
\frac{
\hat{y}_{\ell}-y_{\min}^{(i)}
}{
y_{\max}^{(i)}-y_{\min}^{(i)}
},
\\
y_{\ell}^{\mathrm{norm}}
&=
100
\frac{
y_{\ell}-y_{\min}^{(i)}
}{
y_{\max}^{(i)}-y_{\min}^{(i)}
}.
\end{align}

Thus, normalization is performed using the range of the matched ground-truth values within each chart, rather than independently for each series.

\paragraph{Mean Absolute Error (MAE).}
For $L$ successfully matched point pairs, the normalized Mean Absolute Error is
\begin{equation}
\mathrm{MAE}
=
\frac{1}{L}
\sum_{\ell=1}^{L}
\left|
\hat{y}_{\ell}^{\mathrm{norm}}
-
y_{\ell}^{\mathrm{norm}}
\right|.
\end{equation}

\paragraph{Root Mean Squared Error (RMSE).}
The normalized Root Mean Squared Error is
\begin{equation}
\mathrm{RMSE}
=
\sqrt{
\frac{1}{L}
\sum_{\ell=1}^{L}
\left(
\hat{y}_{\ell}^{\mathrm{norm}}
-
y_{\ell}^{\mathrm{norm}}
\right)^2
}.
\end{equation}

If normalization cannot be performed, the evaluator first uses relative errors,
\begin{equation}
e_\ell^{\mathrm{rel}}
=
100
\frac{
|\hat{y}_\ell-y_\ell|
}{
|y_\ell|
},
\end{equation}
when $|y_\ell|>10^{-6}$. Relative errors are clipped at 100. If neither normalized nor relative errors are available, bounded absolute errors are used:
\begin{equation}
e_\ell^{\mathrm{abs}}
=
\min\left(
|\hat{y}_\ell-y_\ell|,
100
\right).
\end{equation}

MAE and RMSE are then computed as the arithmetic mean and root mean square of the selected pointwise errors, respectively. If either resulting value exceeds 100, both values are discarded and the chart is treated as unparseable.

\paragraph{Aggregation.}
At the chart level, completeness, Point Coverage, MAE, and RMSE are averaged equally across aligned charts. For model-level reporting, these chart-level statistics are therefore aggregated using the number of aligned charts as the weight. Structural metrics such as RFV and CAR are computed directly from the corresponding compound- and chart-level counts.

\subsection{Overall reconstruction performance}

\setcounter{table}{0}
\renewcommand{\thetable}{S\arabic{table}}
\begin{table}[H]
\centering
\caption{Overall reconstruction performance comparison across different models and subplot counts.}
\label{tab:main_results}
\small
\begin{tabular}{lccccccc}
\toprule
\textbf{Model} & \textbf{Subplots} & \textbf{RFV} & \textbf{EC} & \textbf{PR} & \textbf{Point Coverage} & \textbf{MAE} & \textbf{RMSE} \\
\midrule
\multirow{4}{*}{doubao} & 1 & 1.0000 & 0.8366 & 0.7629 & 0.5331 & 8.4932 & 11.9111 \\
                        & 3 & 0.9914 & 0.7817 & 0.7652 & 0.4844 & 9.1226 & 13.1354 \\
                        & 6 & 0.9310 & 0.6887 & 0.7500 & 0.4411 & 11.4940 & 16.2040 \\
                        & 9 & 0.8182 & 0.5754 & 0.7213 & 0.3896 & 12.8941 & 17.5449 \\
\midrule
\multirow{4}{*}{gemini} & 1 & 1.0000 & 0.8731 & 0.8428 & 0.6052 & 5.5119 & 8.1466 \\
                        & 3 & 0.9310 & 0.7771 & 0.8889 & 0.5559 & 8.2033 & 11.6073 \\
                        & 6 & 0.8879 & 0.7373 & 0.8819 & 0.5448 & 7.5849 & 11.0958 \\
                        & 9 & 0.7273 & 0.6860 & 0.8770 & 0.5503 & 8.5136 & 12.0220 \\
\midrule
\multirow{4}{*}{glm}   & 1 & 1.0000 & 0.8343 & 0.8779 & 0.5622 & 8.7179 & 12.6893 \\
                        & 3 & 0.8485 & 0.6727 & 0.8095 & 0.4527 & 11.5136 & 15.9457 \\
                        & 6 & 0.6814 & 0.5442 & 0.7381 & 0.3878 & 14.2203 & 19.7835 \\
                        & 9 & 0.5526 & 0.4872 & 0.7593 & 0.4085 & 16.1116 & 21.2302 \\
\midrule
\multirow{4}{*}{kimi}  & 1 & 1.0000 & 0.8793 & 0.7830 & 0.5586 & 7.9208 & 11.2605 \\
                        & 3 & 0.9957 & 0.8436 & 0.8188 & 0.5426 & 8.4153 & 12.2063 \\
                        & 6 & 1.0000 & 0.8410 & 0.8319 & 0.5364 & 8.7669 & 12.7310 \\
                        & 9 & 0.9870 & 0.8250 & 0.8333 & 0.5285 & 9.4399 & 13.5002 \\
\midrule
\multirow{4}{*}{qwen}  & 1 & 1.0000 & 0.8442 & 0.8403 & 0.5801 & 7.0583 & 10.2186 \\
                        & 3 & 0.7241 & 0.5718 & 0.8214 & 0.4854 & 9.9257 & 14.0155 \\
                        & 6 & 0.6207 & 0.4583 & 0.7593 & 0.3935 & 8.9965 & 12.5950 \\
                        & 9 & 0.9740 & 0.7161 & 0.7793 & 0.3987 & 12.7767 & 17.7086 \\
\bottomrule
\end{tabular}
\end{table}

\subsection{RMSE density figures}
\setcounter{figure}{0}
\renewcommand{\thefigure}{S\arabic{figure}}
\begin{figure}[H]
\centering
\begin{subfigure}[t]{0.4\linewidth}
\centering
\includegraphics[width=\linewidth]{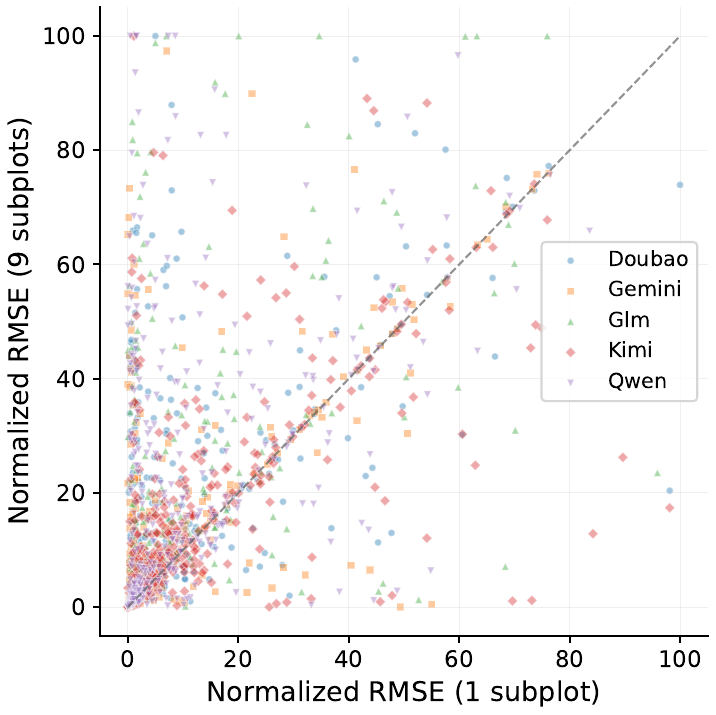}
\caption{}
\label{fig:rmse_1vs9}
\end{subfigure}
\hfill
\begin{subfigure}[t]{0.58\linewidth}
\centering
\includegraphics[width=\linewidth]{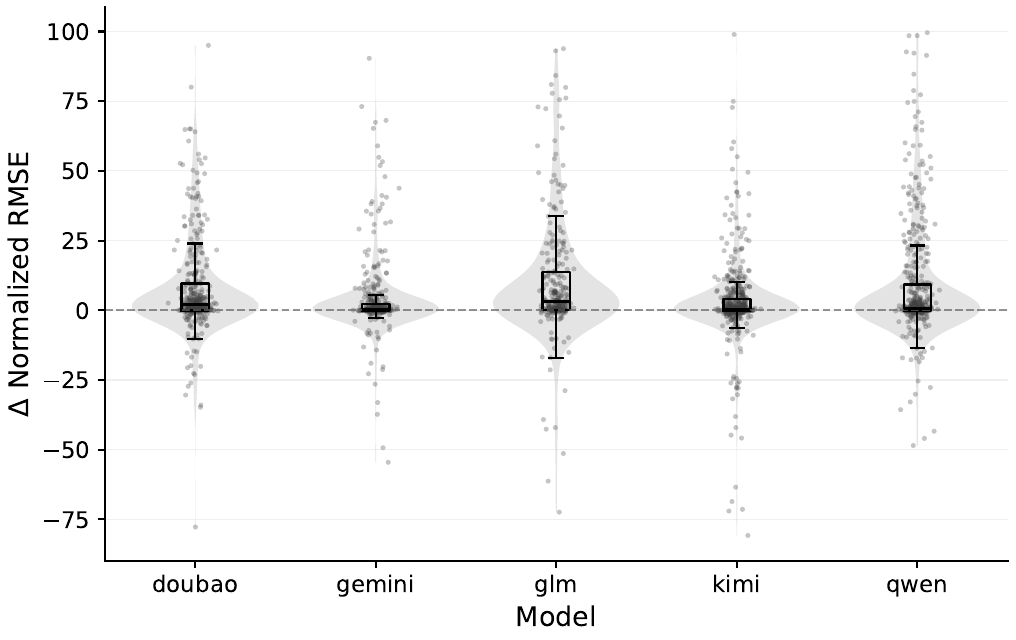}
\caption{}
\label{fig:rmse_delta}
\end{subfigure}
\caption{
Paired chart-level comparison between the 1- and 9-chart conditions.
Each point corresponds to the same source chart; values above the diagonal
indicate higher error under density 9. The right panel shows the distribution
of per-chart error changes.
}
\label{fig:per_chart_analysis_rmse}
\end{figure}

\subsection{Mixed-effects model results}
\begin{table}[H]
\centering
\caption{Regression results for the linear mixed-effects model. The reference categories are kimi for model and 1 for density.}
\label{tab:mixed_effects}
\small
\begin{tabular}{lcccccc}
\toprule
\textbf{Term} & \textbf{Coef.} & \textbf{Std. Error} & \textbf{$z$} & \textbf{$p$-value} & \textbf{CI Low} & \textbf{CI High} \\
\midrule
Intercept & 8.7545 & 0.6280 & 13.9398 & $<$0.0001 & 7.5235 & 9.9854 \\
\midrule
\multicolumn{7}{l}{\textit{Model (ref. = kimi)}} \\
\quad doubao & 0.4171 & 0.6438 & 0.6479 & 0.5171 & -0.8448 & 1.6789 \\
\quad gemini & -2.4025 & 0.6306 & -3.8097 & 0.0001 & -3.6385 & -1.1664 \\
\quad glm & 0.3972 & 0.6235 & 0.6370 & 0.5241 & -0.8249 & 1.6193 \\
\quad qwen & -1.0514 & 0.6309 & -1.6665 & 0.0956 & -2.2879 & 0.1852 \\
\midrule
\multicolumn{7}{l}{\textit{Density (ref. = 1)}} \\
\quad density = 3 & 0.3549 & 0.6339 & 0.5599 & 0.5755 & -0.8876 & 1.5975 \\
\quad density = 6 & 0.6518 & 0.6305 & 1.0338 & 0.3012 & -0.5839 & 1.8875 \\
\quad density = 9 & 1.5519 & 0.6331 & 2.4511 & 0.0142 & 0.3109 & 2.7929 \\
\midrule
\multicolumn{7}{l}{\textit{Model $\times$ Density interactions}} \\
\quad doubao $\times$ 3 & 0.7319 & 0.9068 & 0.8071 & 0.4196 & -1.0454 & 2.5093 \\
\quad gemini $\times$ 3 & 1.9301 & 0.8878 & 2.1741 & 0.0297 & 0.1901 & 3.6701 \\
\quad glm $\times$ 3 & 2.7341 & 0.9067 & 3.0156 & 0.0026 & 0.9571 & 4.5111 \\
\quad qwen $\times$ 3 & 3.0651 & 0.9322 & 3.2881 & 0.0010 & 1.2381 & 4.8922 \\
\quad doubao $\times$ 6 & 3.1270 & 0.9151 & 3.4169 & 0.0006 & 1.3333 & 4.9207 \\
\quad gemini $\times$ 6 & 1.7784 & 0.8924 & 1.9929 & 0.0463 & 0.0294 & 3.5275 \\
\quad glm $\times$ 6 & 4.0658 & 0.9569 & 4.2491 & $<$0.0001 & 2.1903 & 5.9412 \\
\quad qwen $\times$ 6 & 1.5839 & 0.9695 & 1.6338 & 0.1023 & -0.3163 & 3.4840 \\
\quad doubao $\times$ 9 & 3.7571 & 0.9413 & 3.9912 & 0.0001 & 1.9121 & 5.6021 \\
\quad gemini $\times$ 9 & 1.6163 & 0.9217 & 1.7536 & 0.0795 & -0.1902 & 3.4227 \\
\quad glm $\times$ 9 & 6.0766 & 0.9922 & 6.1243 & $<$0.0001 & 4.1319 & 8.0213 \\
\quad qwen $\times$ 9 & 4.8029 & 0.8987 & 5.3440 & $<$0.0001 & 3.0414 & 6.5644 \\
\bottomrule
\end{tabular}
\end{table}
\end{document}